%% file: GlobalSIP_arxiv.tex
\documentclass[10pt]{article}
\usepackage[utf8x]{inputenc}
\usepackage{graphicx}
\usepackage{spconf-arxiv,amsmath,url,amssymb,amsfonts,algorithm,algorithmic,epsfig,subfigure,epstopdf,color, float, wrapfig}
\usepackage[center,font = footnotesize]{caption}
\usepackage{amsthm, url, graphicx, caption, cite, bm, hyperref}
\usepackage{setspace}
\input{macros}

\DeclareMathOperator*{\argmin}{argmin}

\newtheorem{theorem}{Theorem}[section]
\newtheorem{proposition}{Proposition}
\newtheorem{definition}{Definition}

\title{Localized Iterative Methods for Interpolation in Graph Structured Data}
\name{Sunil K Narang, Akshay Gadde, Eduard Sanou and Antonio Ortega \thanks{This work was supported in part by NSF under grant CCF-1018977.}}
\address{Ming Hsieh Department of Electrical Engineering\\
         University of Southern California\\
         kumarsun@usc.edu, agadde@usc.edu, eduardsanou@gmail.com, ortega@sipi.usc.edu}
         
\begin{document}
\maketitle
\begin{abstract}
In this paper, we present two localized graph filtering based methods 
for interpolating graph 
signals defined on the vertices of arbitrary graphs from only a 
partial set of samples. The first method is an extension of 
previous work on reconstructing bandlimited graph signals 
from partially observed samples. The iterative graph filtering 
approach very closely approximates the solution proposed in 
the that work, 
while being computationally more efficient. As an alternative, 
we propose 
a regularization based framework in which 
we define the cost of reconstruction to be a 
combination of smoothness 
of the graph signal and the 
reconstruction error with respect to 
the known samples, and find solutions 
that minimize this cost.   
We provide both a closed form solution and a 
computationally efficient iterative solution 
of the optimization problem. The experimental results 
on the recommendation system datasets 
demonstrate effectiveness of the proposed methods.
\end{abstract}

\section{Introduction}
The field of {\em graph signal processing} 
extends signal processing tools 
designed for regularly sampled 
signals 
to graph datasets~\cite{Shuman_SPM}. 
In the graph representation, the data points 
are represented as nodes connected to 
each other via links. The weights of the links 
usually represent similarity 
between the data points. Each node
stores a sample, and the collection of these samples 
is referred to as a {\em graph signal}. 
In this paper
we discuss an important problem, namely that of 
interpolation 
of missing values from known samples, which   
appears in various applications, such
as matrix/vector completion, 
sampling of high-dimensional 
data, semi-supervised learning etc.
Inspired by standard signal processing 
approaches,  we formulate 
the data interpolation problem as 
a signal reconstruction problem
on a graph. 
This is an extension of our 
previous work in~\cite{SunilAkshay'13}, 
where we used 
sampling results in 
graphs 
to find classes 
of bandlimited (BL) graph
signals that can be reconstructed 
from their partially 
observed samples. 
A class of BL graph signals 
is specified by the 
cut-off graph frequency $\omega$ (denoted as $\omega$-BL), and 
the 
interpolated signal 
is obtained by 
projecting the input signal 
onto the appropriate $\omega$-BL 
subspace 
using a {\em least square} method. The value of 
$\omega$ is estimated using the topology 
of the underlying graph 
and location of known samples in the graph. 
The method proposed in~\cite{SunilAkshay'13} 
provides 
exact reconstruction of $\omega$-BL 
graph signals 
and the best approximation 
(in the least square sense) of arbitrary signals as
$\omega$-BL graph signals.

However, this method of reconstruction 
is computationally expensive 
for large graphs as 
it involves eigenvalue 
decomposition of Laplacian matrix, followed by 
inverse of a square matrix of the size of the graph.
Therefore, in this paper we formulate 
the interpolation problem on graph as an iterative 
graph filtering problem, where the graph filter is designed as an 
ideal low-pass graph filter 
with cut-off frequency $\omega$ as computed
in~\cite{SunilAkshay'13}. 
The proposed iterative 
algorithm is faster and converges to the 
least square reconstruction method in ~\cite{SunilAkshay'13}. 
Also, 
to avoid eigenvalue decomposition of Laplacian matrix, the ideal 
low pass filter 
is approximated with a polynomial of the Laplacian matrix, 
which can be computed efficiently as matrix-vector product without 
the need of eigenvalue decomposition. 

Further, the estimated cut-off frequency $\omega$ is 
only an estimate, and the actual signal 
may not be $\omega$-BL. Therefore, we set up 
a regularized cost that exploits 
the trade-off 
between signal smoothness, and
the reconstruction errors at the 
known samples. The proposed cost function
is based on 
the  data fitting error at the 
known samples and the energy
of the reconstructed signal 
outside the $\omega$-BL subspace. 
The solution of the regularization is computed 
first as an exact solution, followed by 
an approximate solution 
based on the iterative graph filtering approach.
The rest of the paper is organized as follows: in Section~\ref{sec:LSR}, we 
briefly explain the interpolation method proposed in~\cite{SunilAkshay'13}, 
in Section~\ref{sec:ILSR}, provide an iterative graph filtering based solution 
of this interpolation method. In Section~\ref{sec:RBM}, we describe a second method 
for graph signal interpolation, based on a regularization framework, in Section~\ref{sec:RC_sys}
we discuss application of proposed method to item-recommendation systems, and compare results 
with respect to existing methods in Section~\ref{sec:experiments}. 

\section{Sampling Theorem for Band-limited Graph Signals}
\label{sec:LSR}
A graph $G = (\Vc,E)$ 
is a collection of 
nodes  $\Vc=\{1,2,...N\}$ connected together 
by set of links
$E = \{(i,j,w_{ij})\}$, $i,j \in \Vc$.
$(i,j,w_{ij})$ denotes the
link between nodes $i$ and $j$ having weight $w_{ij}$. 
The adjacency matrix $\Wm$ of the 
graph is an $N \times N$ matrix 
such that $W(i,j) = w_{ij}$. The degree
$d_i$ of node $i$ is the
sum of link-weights connected to node $i$.
The degree matrix $\Dm = diag\{d_1,d_2,...,d_N\}$ 
is a diagonal 
matrix. The combinatorial Laplacian matrix 
is defined as $\Lm = \Dm - \Wm$. 
The corresponding symmetric 
normalized Laplacian matrix is
$ \Lcb= \Dm^{-1/2}\Lm \Dm^{-1/2}$. We use the normalized Laplacian matrix because it is closely related to the random walk matrix and is shown to produce superior classification results~\cite{Zhou04}.
We consider only undirected graphs without 
self loops for which $\Lcb$ is a 
symmetric positive semi-definite matrix. 
Therefore, it has the eigenvalue decomposition:
\begin{equation}
\Lcb = \Um\Lambda \Um^t = \sum_{i=1}^N \lambda_i \uv_i \uv_i^t,
\label{eq:EVD}
\end{equation}
with a diagonal eigenvalue matrix $\Lambdam$ 
containing non-negative eigenvalues 
$\{ \lambda_1, \lambda_2 \dots \lambda_N \}$ arranged in a non-decreasing order 
at the diagonal, 
and a unitary matrix $U$ containing corresponding eigenvectors $\uv_i$. 
A graph signal is a function $f:\mathcal{V}\rightarrow \mathbb{R}$ defined on the vertices of the graph. It can be represented as a vector $\fv \in \mathbb{R}^N$ where the $i$th component represent the function value on the $i$th vertex.
Eigenvectors and eigenvalues of $\Lcb$ are used 
to define Fourier transform for graph signals~\cite{Hammond'11,Shuman_SPM,Pesenson'08}. 
Eigenvalues  $\lambda_i$
are the graph frequencies which are 
always in the range $[0,2]$, 
and eigenvectors serve as the 
corresponding basis vectors. 
Every graph signal can be represented with basis 
$\Um$ as $\fv = \sum_i \tilde{f}(\lambda_i)\uv_i$, where $\tilde{f}(\lambda_i) = \langle \fv,\uv_i \rangle$ is the {\em graph Fourier transform} (GFT) of $\fv$.


In classical signal processing, the signal being 
bandlimited implies that the energy of the signal 
is zero above a certain frequency. 
%
The spectral analysis of graph signals offers a
similar interpretation. Following definitions and results 
were used in~\cite{SunilAkshay'13}
to design reconstruction algorithm 
for graph signals.
\begin{definition}[Band-limited graph signal ~\cite{Pesenson'08}]
A signal on a graph $G$ is said to be band-limited to the graph frequency band $[0,\omega)$, if its GFT has support only at frequencies $[0,\omega)$.
\end{definition}
The space of $\omega$-bandlimited signals is called Paley-Wiener space and is given by 
\begin{equation}
PW_{\omega}(G) = \{\fv: \; \tilde{f}(\lambda) = 0 \quad \text{if} \quad \lambda \geq \omega \}
\end{equation}
\begin{definition}[$\Lambda$-set]
A set $\Qc \in \Vc$ is a $\Lambda$-set if all graph signals $\boldsymbol{\phi}$ with support on $\Qc$ (i.e. $\boldsymbol{\phi}(v) = 0$ if $v \notin \Qc$) satisfy
\begin{equation}
\|\boldsymbol{\phi}\| \leq \Lambda \|\Lcb \boldsymbol{\phi} \| \quad \ldots (\Lambda > 0)
\end{equation}
\end{definition}
\begin{theorem}[Sampling theorem~\cite{Pesenson'08}]
All graph signals $\fv \in PW_{\omega}(G)$ can be uniquely 
recovered from a subset of its samples on $\Sc$ if $\Sc^c = \Vc - \Sc$ is a $\Lambda$-set such that $0 < \omega < 1/\Lambda$.
\label{thm:sampling_thm}
\end{theorem}
The following result~\cite{SunilAkshay'13} computes the maximum $\omega$ such that any signal 
in $PW_{\omega}(G)$ can be reconstructed given a subset of known samples $\Sc$ on any graph $G$.
\begin{proposition}[Cut-off frequency~\cite{SunilAkshay'13}]
Let $(\Lcb^2)_{\mathcal{S}^c}$ be the submatrix of $\Lcb^2$  containing 
only the rows and columns corresponding to unknown set $\mathcal{S}^c$.
Let $ \sigma^2_{min}$ to be the smallest eigenvalue of $(\Lcb^2)_{\mathcal{S}^c}$.
Any $\fv \in PW_{\omega}(G)$ with $\omega = \sigma_{min}$ can be uniquely recovered from its samples on $\Sc$.
\label{prop:omega_max}
\end{proposition}

\subsection{Least Squares Reconstruction}
Proposition~\ref{prop:omega_max} gives a condition on the GFT of a graph signal such that unique reconstruction is possible from its given known subset of samples. A simple way to do this reconstruction is to solve a least-squares problem in the spectral domain as explained below.

Let $\lambda_k$ be the largest eigenvalue of $\Lcb$ less than $\omega$.
An $\omega$-bandlimited signal can be written (under appropriate permutation) as
\begin{equation}
\begin{bmatrix}
\fv(\Sc)\\
\fv(\Sc^c)
\end{bmatrix} 
= 
\begin{bmatrix}
\uv_1(\Sc) & \uv_2(\Sc) & \cdots & \uv_k(\Sc) \\
\uv_1(\Sc^c) & \uv_2(\Sc^c) & \cdots & \uv_k(\Sc^c)
\end{bmatrix}
\begin{bmatrix}
\alpha_1\\
\alpha_2\\
\vdots\\
\alpha_k
\end{bmatrix}\\
\end{equation}
Let $\boldsymbol{\alpha} = [\alpha_1,\alpha_2,\ldots,\alpha_k]^t$ and
\[
\begin{bmatrix}
\uv_1(\Sc) & \uv_2(\Sc) & \cdots & \uv_k(\Sc) \\
\uv_1(\Sc^c) & \uv_2(\Sc^c) & \cdots & \uv_k(\Sc^c)
\end{bmatrix}
=
\begin{bmatrix}
(\Um_k)_{\Sc}\\
(\Um_k)_{\Sc^c}
\end{bmatrix}
\]
$\boldsymbol{\alpha}$ can be obtained by calculating a least squares solution to $\fv(\Sc)= (\Um_k)_{\Sc} \boldsymbol{\alpha}$. Then, 
the unknown signal values are given by
\begin{equation}
\fv(\Sc^c) = (\Um_k)_{\Sc^c}\left((\Um_k)_{\Sc}^t (\Um_k)_{\Sc}\right)^{-1} (\Um_k)_{\Sc}^t \fv(\Sc)
\label{eq:LS_sol}
\end{equation}
The sampling theorem guarantees that the there exist a unique
solution to the above least squares problem, which
is equal to the original signal 
$\fv$ if $\fv \in PW_{\omega}(G)$.
On the other hand, if $\fv \notin PW_{\omega}(G)$,
we still get a unique  
least square approximation 
of $\fv$ in  $PW_{\omega}(G)$ space. 
The choice of the cut-off 
frequency $\omega$  (estimated 
from Theorem~\ref{thm:sampling_thm})
is still crucial, even though
the reconstructed signal in this case
may not be the best solution 
in terms of 
reconstruction errors. 
This is because, for a 
frequency $\omega'$ higher than 
$\omega$, there exists a LS solution but
the sampling theorem guarantee fails. This means 
that there may be infinitely many LS solutions 
in the $PW_{\omega'}(G) \supset PW_{\omega}(G)$ space, each 
giving a different 
interpolation result
at the unknown samples. 
Therefore, in~\cite{SunilAkshay'13} we used $\omega$ 
as the cut-off frequency 
for all reconstructed graph signals. The proposed method in~\cite{SunilAkshay'13} 
provides good interpolation results when applied to item-recommendation problem. However, the algorithm 
is computationally expensive as it requires computation of eigenvalues of the Laplacian matrix. In the next section, we provide 
an iterative method for solving the above reconstruction problem.

\section{Iterative Least Square Reconstruction}
\label{sec:ILSR}
Our proposed method is similar to the 
Papoulis-Gerchberg algorithm~\cite{Papoulis,Gerchberg,Sauer} 
in classical signal processing which is used to reconstruct 
a band-limited signal from irregular samples. 
It is a special case of projection 
onto convex sets (POCS)~\cite{Youla}, where 
the convex sets of interest in this case are:
\begin{align}
C_1 = \{\xv: \Jm\xv = \Jm\fv \} \\
C_2 = PW_{\omega}(G)
\end{align}
Here $\Jm:\mathbb{R}^N \rightarrow \mathbb{R}^M$ denotes the downsampling operator where $M$ is the size of the known subset $\Sc$ of samples.
At $k^{th}$ iteration, the solution
$\fv_k$ is obtained from $\fv_{k-1}$, and satisfies the following two constraints:
(1)~the signal equals the known values on the sampling set (i.e., $\fv_k \in C_1$).
(2)~the signal is $\omega$-bandlimited, 
where $\omega$ is computed using Proposition~\ref{prop:omega_max} (i.e., $\fv_k \in C_2$).
We define $\Pm:\mathbb{R}^N \rightarrow PW_{\omega}(G)$ to be the low-pass graph filter such that
\begin{equation}
\yv = \Pm \xv \Rightarrow \yv \in PW_{\omega}(G)
\end{equation}
$\Pm$ can be written in graph spectral domain as $\Pm = \Hm(\Lcb) = \sum_{i = 1}^N h(\lambda_i) \uv_i \uv_i^t$ where
\begin{equation}
h(\lambda) = \left\{ 
  \begin{array}{l l}
    1 & \quad \text{if $\lambda < \omega$}\\
    0 & \quad \text{if $\lambda \geq \omega$}
  \end{array} \right.
  \label{eq:ideal_kernel}
\end{equation}
We define the downsample then upsample (DU) operation as 
\begin{equation}
\fv_{du} = \Jm^t\Jm \fv \Rightarrow  \fv_{du}(\Sc) = \fv(\Sc) \text{ and }\fv_{du}(\Sc^c) = \mathbf{0}.
\end{equation}
With this notation the proposed iterative algorithm can be written as
\begin{align}
\fv_0 &= \Pm\fv_{du} \nonumber\\
\fv_{k+1} &= \Pm(\fv_k + \Jm^t\Jm(\fv_{du} - \fv_k))
\label{eq:iter_band}
\end{align} 
At each iteration the algorithm resets the signal samples on $\Sc$ to the actual given samples and then projects the signal onto the low-pass space $PW_{\omega}(G)$. 

\subsection{Convergence}

We define the operators $\Bm:\mathbb{R}^N \rightarrow \mathbb{R}^N$ and $\Tm:C_2\rightarrow C_2$ corresponding to iteration in~\eqref{eq:iter_band} as
\begin{align}
\Bm\xv &= \xv + \Jm^t\Jm(\fv_{du} - \xv)\\
\Tm\xv &= \Pm(\xv + \Jm^t\Jm(\fv_{du} - \xv)) = \Pm\Bm \xv
\end{align}
It has been shown~\cite{Youla} that an iterative algorithm of the form $\xv_{k+1} = \Tm\xv_k$ converges to a fixed point of $\Tm$ if 
\begin{enumerate}
\item $\Tm$ is non-expansive, i.e., $\|\Tm\xv - \Tm\yv\| \leq \|\xv - \yv\|$
\item $\Tm$ is asymptotically regular, i.e., $\|\Tm \xv_{k+1} - \Tm \xv_{k}\| \rightarrow 0$ as $k \rightarrow \infty$.
\end{enumerate} 
$\Pm$ is a bandlimiting operator and hence is non-expansive. $\Bm$ is non expansive because $\|\Bm\xv - \Bm \yv\| = \|(\mathbf{I}-\Jm^t\Jm) (\xv - \yv)\| \leq \|\xv - \yv \|$.
Since both $\Pm$ and $\Bm$ are non-expansive, $\Tm$ is also non-expansive. Asymptotic regularity of $\Tm$ can also be proved as shown in~\cite{Sauer}.
Note that if $\fv$ is a fixed point of $\Tm$ then $\fv \in C_1 \cap C_2$. From the sampling theorem $\fv \in C_1 \cap C_2$ is unique. 
So the asymptotic solution of the proposed algorithm converges to the solution of the least square projection method described in previous section.
 
\subsection{Iterative reconstruction with polynomial low pass filter}
The low pass filter $\Pm$ above is a spectral 
graph filter with an ideal  brick wall type spectral response. Thus, the exact 
computation of $\Pm$ requires eigenvalue 
decomposition of the Laplacian matrix, which is computationally very expensive for large matrices. 
However, it 
is possible to approximate the ideal filtering operation as a matrix 
polynomial in terms of $\Lcb$, that can be implemented efficiently using only matrix vector products. 
Thus we replace 
$\Pm$ in~(\ref{eq:iter_band}) with an approximate low pass filter $\Pm^{poly}$ given as:
\begin{equation}
\Pm^{\text{poly}} = \sum_{i = 1}^N  \left(\sum_{j = 0}^k a_j \lambda_i ^j \right) \uv_i \uv_i^t = \sum_{j = 0}^k a_j \Lcb ^j
\end{equation}
We specifically use 
the truncated Chebychev polynomial expansion of any spectral kernel $h(\lambda)$, 
as proposed in
~\cite{Hammond'11}, in our experiments.
The proposed iterative least square method with polynomial low pass filter,
is termed as iterative least square (ILSR) in this paper. 

%

\section{Interpolation Based on Regularization}
\label{sec:RBM}
The method presented above does not allow solutions 
from outside the $PW_{\omega}(G)$ space. This
is advantageous if 
the input signal 
belongs to or is close to the subspace spanned by  
$\omega$-BL signals. In general, 
for real world datasets 
such as recommendation systems, 
the graph signals tends to be 
smooth but not  
exactly band-limited. 
Therefore, we use a 
graph regularization framework in which 
we set up the 
cost of reconstruction as:
\begin{equation}
\fv^* = \argmin_{\xv} \underbrace{\|\Jm(\fv_{du} - \xv)\|^2}_{A} + \alpha \underbrace{\|\Hm\xv\|^2}_{B}
\label{eq:regularized_cost}
\end{equation}
where $A$ is the {\em data-fitting} term which computes the 
error between reconstructed signal and the original signal at the known samples and 
$B$ is the Euclidean norm of the output of a highpass graph filter $\Hm$. 
Thus, the term $A$ in the cost function penalizes the 
signals that are different from original signal at the known nodes, 
and the term $B$ penalizes signals that have 
significant high frequency components.
{\bf Note that the optimal solution of~(\ref{eq:regularized_cost}) converges 
to the least square solution computed in~(\ref{eq:iter_band}), 
if $\Hm = \Id - \Pm$ and $\alpha \to \infty$.} In our experiments,  $\Hm$ is 
a chosen as a spectral graph 
transform with spectral kernel 
$h(\lambda) = \exp(-1/\lambda)$. 
The problem in~(\ref{eq:regularized_cost}) has a well known 
closed form solution given as:
\begin{equation}
\fv^* = (\Jm^t \Jm + \alpha \Hm^t\Hm)^{-1}\Jm^t\Jm\fv_{du} = (\Jm^t \Jm + \alpha \Hm^t\Hm)^{-1}\fv_{du}
\label{eq:regularization_closed_form}
\end{equation}
However, a direct implementation is computationally expensive, as it involves 
both the eigenvalue decomposition of the Laplacian matrix (to compute highpass filter $\Hm$) and 
inversion of a graph size matrix. Therefore, we propose an approximate 
iterative solution of the optimization problem 
in~(\ref{eq:regularized_cost}), similar to the method based on POCS~\cite{Katsaggelos} 
described in Section~\ref{sec:ILSR}.
\begin{align}
\fv_0 &=\fv_{du} \nonumber\\
\fv_{k+1} &= (\mathbf{I} - \beta \alpha \Hm^t\Hm)\fv_k + \beta \Jm^t\Jm (\fv_{du}-\fv_k)
\label{eq:iter_reg}
\end{align}
The parameter $\beta$ is chosen to ensure convergence and maximize the rate of convergence. Replacing  the spectral transform $\Hm^t\Hm$ by its polynomial approximation, we get a local iterative method for regularized graph signal recovery. Since we use a continuous function of $\lambda$ to construct the regularization term, even a low degree polynomial approximation does not greatly affect the solution.

\section{Application: Recommendation Systems}
\label{sec:RC_sys}
We apply the proposed interpolation method for 
collaborative filtering 
in recommendation systems. The input 
in this problem
is a partially observed user-item rating matrix $\Rm$, such that $\Rm(u,m)$ 
is the rating given by user $u$ to the item $m$. Based on 
this information, 
the system 
predicts new user-movie ratings. Following the setup in~\cite{SunilAkshay'13}, 
an item-item graph $G_0$ is computed 
using partially observed rating matrix $\Rm$. The weight of the link between 
each pair of items $i$ and $j$ is computed as the
cosine similarity~\cite{sarwar_kNN} between $i$ and $j$ 
based on the training samples. 
For each test user $u$, 
we define $\Sc$ to be the set of 
items with known ratings, 
and
and $\Uc$ to be the set of test items. 
We compute the subgraph 
$G_u = (\Sc \cup \Uc, E_u)$ of $G_0$, corresponding to  
the subset $\Sc \cup \Uc$ of nodes. 
We define DU signal 
$\fv_u$ for $u$ to be of size 
$|\Uc \cup \Sc|$, with $\fv_u(\Uc)= 0$ and $\fv_u(\Sc)$ 
equal to known ratings. Subsequently, we compute interpolated signal $\hat \fv_u$ 
by using graph based interpolation. 
\subsection{Graph Simplification}
The item-item graphs computed using cosine similarity (as above),
usually end up being highly connected if the rating matrix $\Rm$ is not sparse. 
The graph frequencies of very dense graphs are not uniformly 
distributed and hence not very informative in describing the smoothness of the signal.  
We observe that simplification of the item-item graph as a {\em $K$ nearest neighbor} (KNN) leads to more uniform  and informative 
distribution of graph frequencies. Therefore, we sparsify the subgraph $G_u$ obtained for user $u$ by 
connecting each item $i$ in $G_u$ 
by at most top $K$ of its known neighbors, (ordered according to the decreasing link weights with item $i$). 
The best value of $K$ is determined 
empirically to be around 
$30$ in this paper. 

\subsection{Bilateral Link-Weight Adjustment}
In addition to the sparsification step, 
the weights of the links between known samples $\Sc$  
in the subgraph $G_u$ are adjusted to reflect the user $u$'s preferences, 
as is done in~\cite{SunilAkshay'13}. 
This adjustment step makes sense 
since subgraph $G_u$ is the result 
of observing average correlation 
over a set of training 
users (multiple instances), 
and the signal $\fv_u$ corresponds to 
a single test user $u$. Specifically, we 
use bilateral-like 
weights for the links between known set of nodes, the exact implementation of which 
can be found in~\cite{SunilAkshay'13}. 
\section{Experiments}
\label{sec:experiments}
In our experiments, we use three different recommendation system datasets to evaluate 
the performance of proposed algorithms. Each dataset contains 
a reduced set of $100k$ randomly 
selected entries of user-item-ratings.  The properties of the datasets 
are given in Table~\ref{tab:datasets}. 
\begin{table}[h]
\centering
  \begin{tabular}{|p{2.5cm}|p{1.2cm}|p{1.2cm}|p{0.8cm}|p{0.82cm}|p{0.82cm}|}
  \hline
    Dataset              & \# users & \# items & rating range & $mod[u]$ & $mod[i]$ \\
    \hline
    Movielens~{\cite{movielens-data}} 	& 943 & 1682 & 1--5 & 215 &57 \\
    Jester~\cite{jester-data}            	& 1412 & 100 & 0--20 &  80 & 1104 \\
    BX-Books~\cite{bx-books-data}          	& 6299 & 7046 & 1--10 &  80 & 5 \\
    \hline
  \end{tabular}
  \caption{Datasets used in the experiments. $mod[u]$ and $mod[i]$: mode number of the ratings per user
  and per item, respectively. The ratings of Jexter datasets are originally fractional values in the range $-10$ to $9$, which are 
rescaled to the range $0$ to $20$ and rounded to integer value.}
\label{tab:datasets}
\end{table}
In each case, we perform a $5$ fold  
cross-validation, in which we split the 
rating entries into $5$ sets of approximately the same size. 
Then we evaluate the dataset $5$ times, 
always using one set for testing and 
all other sets for training. 
In each iteration, 
an item-graph is formed 
from the training samples, 
as described in 
Section~\ref{sec:RC_sys}. 
\begin{table}[t]
\centering
  \begin{tabular}{|c|c|c|c|c|c|c|}
  \hline
    Dataset              & KNN & PMF & RBM & IRBM & LSR & ILSR  \\
    \hline
    Movielens 	  	& 0.2482  &  0.2513  &  {\bf 0.2415}  &   0.2450  &  0.2514  &  0.2466 \\
    Jester            	& 0.2348  &  {\bf 0.2299}  &  0.2304  &  0.2341  &  0.2344  &  0.2315 \\
    BX-Books          	& 0.2677  &  0.2093  &  {\bf 0.1966}  &  0.2138  &  0.2651  &  0.2828 \\
    \hline
  \end{tabular}
  \caption{Normalized RMSE results of the algorithms applied to the different datasets. 
   KNN: $K$ nearest neighbor method, PMF: probabilistic 
  factorization method, RBM: Regularization based method, IRBM: Iterative regularization based method, 
  LSR: Least Square Reconstruction, ILSR: iterative least square reconstruction.}
\label{tab:rmse}
\end{table}
The accuracy of the proposed methods depends 
to a large extent on the accuracy 
of computing link weights 
between items. 
Comparing
the three databases in Table~\ref{tab:datasets}, 
the Jester database contains ratings of only
$100$ items (jokes). This is 
also the dataset with the highest votes 
per item. Therefore, 
we expect the link weights in the
item graph, 
as computed from the training data  
to be highly accurate. 
On the contrary, the 
books database has the smallest number 
of ratings per item, which means that the 
weight of the links in 
the item-graph
may be noisy and not very accurate. 
The movielens dataset 
seems to have enough ratings 
per item to properly compute 
the weights. Note that, insufficient training ratings 
is common problem in all collaborative filtering 
methods.
Further, the accuracy of the proposed methods also
depends on the number of movies rated 
by each user. 
In all the above databases, 
each user ranked enough items to 
give us a good prediction. 
Table~\ref{tab:rmse} shows the 
RMSE of the proposed methods 
with some of the existing methods. 
To fairly compare the performance, the 
actual RMSE obtained for each dataset
is normalized to be  between $0$  and  $1$, 
by dividing 
it with the maximum possible error 
(i.e., maximum rating - minimum rating).
The best RMSE obtained in each dataset 
is represented with bold letters. It can be seen that 
the proposed regularized based 
kernel method (RBM) performs the best
in the MovieLens and books dataset, 
and very close to the best 
method (PMF) in the Jester dataset. 
The iterative approximation 
of RBM (i.e., IRBM) also performs very close 
to the RBM method. However, 
in case of least square methods the iterative 
algorithm (ILSR) performs better than the 
exact method(LSR), on 
the movie and jokes datasets. This is because the ILSR method
uses approximate low-pass filters which allow
some energy to be in the frequencies bands higher than the cutoff $\omega$, 
and is therefore closer to the RBM method.
\section{Conclusions}
\label{sec:conlcusions}
In this paper, we presented two  localized iterative graph filtering based methods 
for interpolation of graph signals from partially observed samples. The methods 
are implemented 
on recommendation system datasets, 
and provide reasonably good results when compared with the existing methods.

\bibliographystyle{IEEEbib}
\bibliography{InterpRefs}

\end{document}

%% file: macros.tex
\usepackage{bm}

\long\def\comment#1{}

\newfont{\bbb}{msbm10 scaled 700}

\newfont{\bb}{msbm10 scaled 1100}

\newcommand{\fv}{{\bf f}}

\newcommand{\uv}{{\bf u}}

\newcommand{\xv}{{\bf x}}
\newcommand{\yv}{{\bf y}}

\newcommand{\Bm}{{\bf B}}

\newcommand{\Dm}{{\bf D}}

\newcommand{\Hm}{{\bf H}}
\newcommand{\Id}{{\bf I}}
\newcommand{\Jm}{{\bf J}}

\newcommand{\Lm}{{\bf L}}

\newcommand{\Pm}{{\bf P}}

\newcommand{\Rm}{{\bf R}}

\newcommand{\Tm}{{\bf T}}
\newcommand{\Um}{{\bf U}}
\newcommand{\Wm}{{\bf W}}

\newcommand{\Qc}{{\cal Q}}

\newcommand{\Sc}{{\cal S}}

\newcommand{\Uc}{{\cal U}}

\newcommand{\Vc}{{\cal V}}

\newcommand{\Lcb}{{\bm {\mathcal L}}}

\newcommand{\Lambdam}{\hbox{\boldmath$\Lambda$}}
